\documentclass[10pt]{iopart}
\usepackage{iopams}
\usepackage[
    backend=biber,
    style=authoryear,    % Use 'authoryear' style
    maxnames=2,           % Max names to show before using "et al."
    minnames=1            % Min names to show before "et al." is allowed
]{biblatex}  
\usepackage[colorlinks=true, linkcolor=blue, urlcolor=blue, citecolor=blue]{hyperref}

\usepackage{algorithm}
\usepackage{algpseudocode}

\usepackage{xcolor} 
\usepackage{graphicx}
\usepackage{bookmark}
\begin{document}

\title[]{Gradient flow of magnetic inclination and intensity as a GPS-free guidance mechanism}

% Nature's invisible roads: how the gradient flow of Earth's magnetic field guides animals to their destination
\author{Zayed Herma$^{1,2}$, Philippe Cardin$^{2}$
and Andréa Vassilev$^{1}$}

\address{$^{1}$ Univ. Grenoble Alpes, CEA, Leti,
38000 Grenoble, France}

\address{$^{2}$ Univ. Grenoble Alpes, CNRS, ISTerre,
38000 Grenoble, France}
% \ead{zayed.herma@cea.fr}
\vspace{10pt}

\begin{abstract}
It is well established that some animals, such as sea turtles, sense the Earth's magnetic field and use it as a primary cue for long-distance navigation across the vast and featureless ocean. This remarkable ability has inspired the development of bio-inspired magnetic navigation (BIMN), which seeks to mimic these animals to develop robust and frugal navigation methods. BIMN exploits the Earth's magnetic field, sometimes described as ``Nature's GPS'', as a natural navigation infrastructure, providing a potential alternative to vulnerable satellite navigation systems. However, unlike geographic coordinates, magnetic coordinates extracted from the Earth's magnetic field do not provide a one-to-one mapping to geographic locations. Different locations can exhibit similar magnetic signatures, creating ambiguity and making BIMN inherently more challenging. Despite this fundamental challenge, we demonstrate that, within predefined regions of the Earth, magnetic inclination and intensity provide sufficient information to navigate to any destination. By formulating BIMN as a gradient flow control and guidance problem, we show that the resulting gradient flow act as current that carries the system and can be controlled to guide it through intermediate waypoints toward the desired destination. The resulting framework provides a GPS-free navigation mechanism with potential applications in both engineering and biology. In engineering, it can be used to establish a geomagnetic navigation infrastructure through which prescribed routes can be defined and followed, while in biology, it can be used to analyse the migration routes of animals known to rely on magnetoreception.
\end{abstract}

\vspace{2pc}
\noindent{\it Keywords}: GPS-free navigation, bioinspired, magnetoreception, geomagnetic field, control, gradient flow
\maketitle
\ioptwocol

\clearpage
\section{Introduction}
\label{sec:introduction}

The Earth’s magnetic field is one of the few consistently available sources of spatial information across the vast, featureless ocean. Sea turtles undertake remarkable long-distance migrations, suggesting that they may rely heavily on geomagnetic cues for navigation \parencite{walker_magnetic_2002, lohmann_there_2019}. It is well established that these animals sense the Earth's magnetic field through magnetoreception \parencite{johnsen_physics_2005}, a sensory ability that humans are not believed to possess \parencite{chae_human_2022}. However, despite lacking this natural ability, humans have long exploited the Earth's magnetic field for navigation. Since the Middle Ages, magnetic needles used as compasses have enabled sailors to navigate across the oceans \parencite{gubbins_four_2000}. More recently, technologies such as magnetometers have enabled artificial sensing of the Earth’s magnetic field, allowing autonomous systems to implement navigation strategies inspired by these species. Such geomagnetic navigation provides a robust alternative to satellite-based navigation in GPS-denied environments, particularly because it is inherently resilient to jamming and spoofing. These characteristics have motivated the development of bio-inspired magnetic navigation (BIMN), which aims to provide robust and frugal navigation techniques using the Earth’s naturally available magnetic field \parencite{zhongwei_zhao_long-distance_2014, qi_bioinspired_2018, taylor_bioinspired_2018, zhang_geomagnetic_2021, zhang_bio-inspired_2021, gill_navigation_2024, herma_bio-inspired_2026}.

\begin{figure*}[t]
\centering
\fbox{%
\begin{minipage}{\dimexpr\textwidth-2\fboxsep-2\fboxrule\relax}
\centering

% Panel (a)
\begin{minipage}[b]{0.48\linewidth}
    \centering
    \includegraphics[width=0.8\linewidth]
    {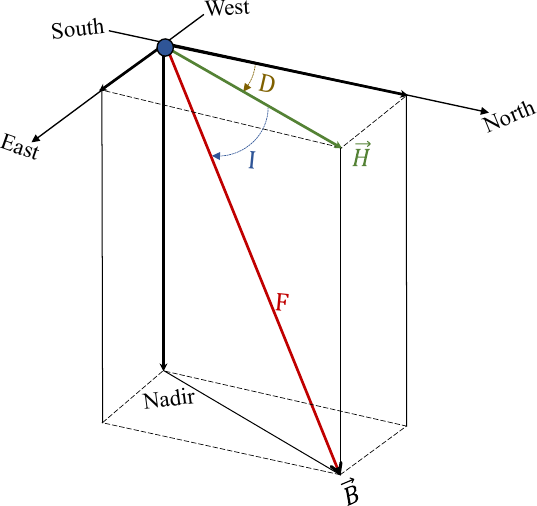}\\[2pt]
    \textbf{(a)}
\end{minipage}
\hfill
% Panel (b)
\begin{minipage}[b]{0.48\linewidth}
    \centering
    \includegraphics[width=0.8\linewidth]
    {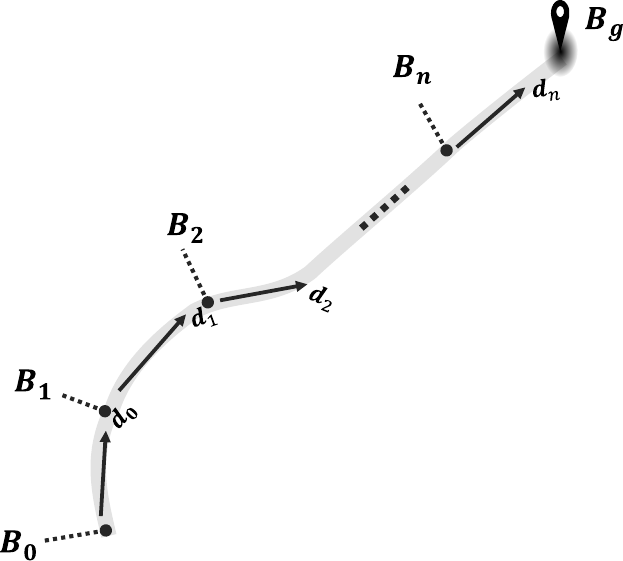}\\[2pt]
    \textbf{(b)}
\end{minipage}

\vspace{1em}

% Panel (c)
\begin{minipage}[b]{0.8\linewidth}
    \centering
    \includegraphics[width=\linewidth]
    {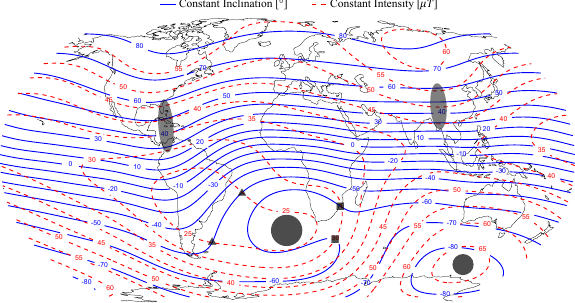}\\[2pt]
    \textbf{(c)}
\end{minipage}

\caption{%
(a) Reproduced from \parencite{herma_bio-inspired_2026}. Illustration of the magnetic field parameters in the (north, east, nadir) reference frame. The inclination ($I$) is the angle between the magnetic field vector $\vec{B}$ and its horizontal projection $\vec{H}$, and the declination ($D$) is the angle between $\vec{H}$ and geographic north. The total field intensity is denoted by $F$. (b) Schematic illustration of the navigation mechanism. The system memorizes the magnetic parameters at the destination (map sense), $\bi{B}_g$, and compares them with those measured at its current position, $\bi{B}_k$, to infer the local direction $\bi{d}_k$ toward the target (compass sense). (c) Illustration that shows examples of the main navigation challenges. The two gray triangles represent locations with identical magnetic parameters, as do the two gray cubes, illustrating the non-uniqueness of magnetic coordinates. The gray circles indicate regions where both magnetic coordinates exhibit nearly vanishing spatial variation, while the gray ellipses indicate regions where they vary in nearly the same direction (parallel isolines). These regions exhibit gradient degeneracy, making navigation toward the destination challenging. Finally, differences in the relative scaling and spatial variation of the magnetic parameters, together with the geometry and curvature of their isolines, can affect the shape and optimality of the navigation path.
}
\label{fig:magnetic_overview}

\end{minipage}%
}
\end{figure*}

\paragraph{}
The main challenge in developing BIMN methods lies in the fact that the mechanisms by which these animals use the Earth's magnetic field for navigation remain incompletely understood. Nevertheless, two complementary mechanisms have been proposed to explain their navigation, known in the literature as the magnetic map sense and the magnetic compass sense, in which the Earth's magnetic field provides directional (compass sense) and positional (map sense) information, much like a global positioning system (GPS) \parencite{johnsen_physics_2005, lohmann_magnetic_2007, goforth_learned_2025}. This is supported by growing evidence that animals likely use magnetic parameters such as intensity, inclination, and occasionally declination, which characterize the Earth's magnetic field (see figure~\ref{fig:magnetic_overview}(a)) and exhibit spatial variations that can be used for navigation \parencite{dennis_evidence_2007, brothers_evidence_2015}. Although no clear consensus has emerged in research on animal navigation regarding the precise interpretation of these two mechanisms \parencite{lohmann_misconceptions_2023}, their translation into engineering applications generally follows a common practical framework, which can be implemented as follows. First, the magnetic parameters associated with a given destination are memorized, providing a practical representation of the magnetic map sense. These parameters form a set of magnetic coordinates that identifies the destination, analogous to the GPS coordinates used to identify a location on Earth. During navigation, the local magnetic parameters are continuously measured, providing the magnetic coordinates of the current location. A BIMN method can then compare these local magnetic coordinates with those of the destination to infer the direction toward it, as illustrated schematically in figure~\ref{fig:magnetic_overview}(b). In this way, the method used to infer this direction provides a representation of the magnetic compass sense. The process can be understood through a simple analogy with GPS. In GPS navigation, the current coordinates are continuously compared with the destination coordinates to determine the direction in which to move. Similarly, a lost person may recognize a familiar landmark from far away and move toward it, progressively reducing the visual distance until reaching it. BIMN follows a similar principle. However, unlike GPS coordinates, magnetic parameters do not uniquely identify a location on Earth. Different locations can exhibit similar magnetic signatures, creating ambiguity and making BIMN more challenging. Beyond this non-uniqueness, two additional difficulties arise. First, even when navigation toward the destination is possible, the resulting trajectory may be far from optimal in terms of path length. Second, some regions of the Earth’s magnetic field exhibit gradient degeneracy, where the spatial gradients of two magnetic parameters become nearly aligned or nearly vanish. For example, nearly parallel spatial gradients provide limited independent directional information, while weak spatial gradients insufficiently constrain the direction of motion \parencite{bostrom_where_2012}. In both cases, the navigation method may fail to identify a suitable direction toward the destination. These challenges are shown in figure~\ref{fig:magnetic_overview}(c) using intensity and inclination as magnetic coordinates. These magnetic parameters are obtained from the International Geomagnetic Reference Field (IGRF) \parencite{beggan_international_2026}, a standard model of the Earth's main magnetic field that is updated every five years using measurements collected worldwide.

\paragraph{}

The three main difficulties (non-uniqueness of magnetic coordinates, unnecessarily long trajectories, and gradient degeneracy of magnetic parameters) raise a fundamental question about the reliability of BIMN. Most proposed methods do not discuss the validity of their approach across different areas. In other words, they do not characterize magnetic regions in which a given BIMN method is suitable for navigation. Moreover, its ability to generalize even within a specific area is not guaranteed. These methods differ mainly in how they infer the local direction of motion, as if correctly identifying this direction were sufficient, without taking the complexity of the magnetic environment into consideration. Simulations are typically limited to a single trajectory between one starting point and one destination. This serves only to illustrate the feasibility of the proposed approach rather than to demonstrate its general applicability, since successful navigation between one pair of locations does not guarantee successful navigation between other locations in the same region. This leads to the central question addressed in this work: How can we characterize the regions and conditions under which BIMN is reliable? In other words, are there regions and navigation strategies for which navigation from any starting point to any destination is possible without encountering the three difficulties described above? 

\paragraph{}

This study is based on the method proposed in \parencite{herma_bio-inspired_2026}, which estimates the gradients of magnetic parameters to capture their local spatial variations. It then uses gradient descent to determine the local direction of motion that minimizes the difference between the measured magnetic coordinates and the corresponding target coordinates. This is analogous to navigating toward a familiar landmark by progressively reducing the visual distance to it. While the original method uses gradient descent for navigation, it does not directly address the three main challenges. For example, the method can become trapped in a local minimum due to gradient degeneracy or converge to another global minimum that does not correspond to the actual destination due to the non-uniqueness of magnetic coordinates. We therefore study the more general structure of the resulting gradient flow, which is the set of trajectories generated by the gradient descent dynamics, to enable reliable navigation throughout predefined regions. We show that, within certain regions separated by well-defined frontiers, it is possible to perform BIMN from any initial location to any desired destination. A similar problem has been considered in \parencite{gill_navigation_2024}, where navigable regions are characterized using local stability analysis around the destination. However, the size and shape of these regions depend on the choice of specific navigation parameters, and local stability does not guarantee navigation between arbitrary locations within these regions. In our case, stability near the destination is naturally guaranteed because, under the gradient descent mechanism, the destination acts as an attractor (a stable equilibrium corresponding to a global minimum). We therefore go beyond this local property and characterize regions in which navigation between arbitrary initial and destination locations can be guaranteed. We formulate navigation as a guidance and control problem, where the gradient flow can be seen as a current carrying the system toward the  destination, which can be controlled and guided by selecting appropriate intermediate destinations. In particular, we prove that a suitable sequence of intermediate destinations always exists, enabling reliable navigation to the final destination (see section~\ref{sec:methods}). A simulation example illustrates this result and shows how intermediate destinations can be used to guide the system toward the final desired destination. Finally, we compare our results with four iconic sea turtle migration trajectories reported in \parencite{putman_marine_2018, laforge_uncovering_2023} to assess whether the regions identified in our work are consistent with observed migration paths known to rely on the Earth's magnetic field for navigation \parencite{lohmann_light-independent_1993,lohmann_magnetic_2007} (see section~\ref{sec:results}).

\paragraph{}
As an important practical consequence, the proposed BIMN framework enables the creation of ``invisible" magnetic routes, defined by sequences of magnetic waypoints that guide the system along prescribed paths, analogous to conventional air routes in which aircraft navigate through sequences of predefined geographic waypoints. In this way, we provide a basis for further developing an alternative navigation infrastructure that does not depend on satellite positioning. Such an approach may become increasingly important as growing space debris and cascading collisions associated with Kessler syndrome threaten satellite infrastructure and the services that rely on it, including navigation \parencite{kessler_collision_1978,mariappan_kesslers_2023, thiele_orbital_2026}.

% \FloatBarrier
\clearpage
\section{Methods}
\label{sec:methods}
This section is organized into three parts. The first is about defining the problem and how the parameters can be extracted, with a focus on the map and compass sense interpretation in the gradient descent method proposed by \parencite{herma_bio-inspired_2026}. The second part focuses on defining the navigation regions and proving the existence of a sequence of intermediate points that allows navigation from any starting position to any destination within the defined regions. In other words, this sequence allows the gradient flow to be directly controlled to guide the system toward the destination. The final part proposes a simple parametric approach to optimize the trajectory, reducing the influence of the relative scaling and variations of the magnetic parameters on its shape. As this study is application-oriented, we focus on two geomagnetic parameters, inclination $I$ and intensity $F$, which can be easily extracted in practice using a magnetometer combined with an accelerometer that provides the horizontal reference, without requiring knowledge of the sensor's global orientation. The proposed analyses can be easily extended to include additional magnetic parameters, such as declination, or other spatially varying physical parameters, such as gravitational field strength, which can provide valuable complementary information.

\subsection{Gradient descent as navigation mechanism}
We denote by $\bi{B}_g=(I_g,F_g)$ the magnetic parameters associated with the destination and by $\bi{B}=(I,F)$ the current magnetic parameters. We consider two-dimensional navigation at a fixed altitude above the Earth's surface, where the horizontal position of the system is denoted by $\bi{x}=(x,y)$. The proposed analysis is independent of the coordinate system used to represent $(x,y)$. The position $\bi{x}$ is unknown to the system; only the current magnetic parameters $\bi{B}$ and the memorized destination parameters $\bi{B}_g$ are known. The current magnetic parameters can be extracted using a three-dimensional magnetometer vector and an accelerometer measuring in the same sensor frame. Assuming that the system moves at a constant velocity, the gravity vector $\bi{g}$ can be measured in the sensor body frame or estimated, thereby providing the vertical direction. The magnetic parameters are then given by
\[
F = \|\bi{B}\| \quad \mathrm{in}\ \mu\mathrm{T}\ \mathrm{(microtesla)},
\]
and
\[
I = 90^\circ -
\arccos\left(
\frac{\bi{g}\cdot\bi{B}}
{ \|\bi{g}\|\,\|\bi{B}\|}
\right)
\quad \mathrm{in\ degrees}.
\] The navigation problem can be formalized as following a path that continuously minimizes the objective function defined as
\begin{equation}
\label{eq:objectif_func}
 V = \|\bi{B}_g-\bi{B}\|^2
= \alpha(I_g-I)^2 + (F_g-F)^2,
\end{equation}
where $\alpha$ is a positive normalization constant that accounts for the different units of inclination (degrees) and intensity (microtesla). For simplicity, we set $\alpha=1$ here. Its utility is discussed in section~\ref{sec:path_optimization}. This formulation implicitly encodes the two mechanisms of map sense and compass sense. Minimizing $V$ to nearly zero is equivalent to reaching the destination position defined by $\bi{B}_g=(I_g,F_g)$, while continuously finding the direction that minimizes it is equivalent to the compass sense. The method proposed in \parencite{herma_bio-inspired_2026} to minimize this quantity in equation~\ref{eq:objectif_func} is based on estimating the gradient, denoted by
\[
\nabla \bi{B} =
\left(
\begin{array}{cc}
\displaystyle \frac{\partial I}{\partial x} &
\displaystyle \frac{\partial I}{\partial y} \\[4pt]
\displaystyle \frac{\partial F}{\partial x} &
\displaystyle \frac{\partial F}{\partial y}
\end{array}
\right).
\] 
The gradient is estimated by comparing the magnetic parameters measured by the system with those predicted from its current estimate. As the system moves, this comparison is used to continuously update the gradient estimate, since the gradient changes from one location to another. In more detail, the change between two consecutive positions during the gradient descent is described by the discrete-time dynamics
\begin{equation}
\label{eq:discrete_dynamics}
\bi{x}_{k+1}
=
\bi{x}_{k}
+
v\,\Delta t\,\bi{d}_k,
\end{equation}
where $v$ is the speed of the system, $\Delta t$ is the sampling period, and $\bi{d}_k$ is the normalized descent direction, defined as
\begin{equation}
\label{eq:descent_direction}
\bi{d}_k
= \frac{-\nabla V_k}{\left\|\nabla V_k\right\|} = 
\frac{
\nabla \bi{B}^{\mathsf T}_{k}
\left(\bi{B}_g-\bi{B}_{k}\right)
}{
\left\|
\nabla \bi{B}^{\mathsf T}_{k}
\left(\bi{B}_g-\bi{B}_{k}\right)
\right\|
}. 
\end{equation}

After the system moves from $\bi{x}_k$ to $\bi{x}_{k+1}$, the magnetic parameters at the new position can be approximated using a first-order expansion:
\[
\bi{B}_{k+1}
\approx
\bi{B}_{k}
+
\nabla \bi{B}_{k}
\left(\bi{x}_{k+1}-\bi{x}_{k}\right).
\]

The difference between this predicted value and the actual measurement of $\bi{B}_{k+1}$ provides information that can be used to refine the gradient estimate. In this way, the gradient is continuously updated as the system moves. The key point in this approach is that the descent direction toward the destination, defined in \ref{eq:descent_direction}, follows the negative gradient of the objective function, $-\nabla V_k$. Thus, $\bi{d}_k$ provides the local direction that reduces the difference between the current magnetic parameters $\bi{B}_{k}$ and those associated with the destination $\bi{B}_{g}$. In addition,  since the measured magnetic parameters are independent of the system's heading, the gradient estimation can accommodate heading errors. If the estimated heading is biased, the gradient is estimated accordingly, so that the resulting navigation direction remains consistent with the magnetic measurements. This makes the gradient descent navigation approach robust to such errors and to external disturbances, such as wind or ocean currents. It also reflects the bio-inspired nature of the approach, as animals can navigate without relying on a highly precise estimate of their heading.

\paragraph{}
We generalize this approach from a dynamical systems perspective. Equation~\ref{eq:discrete_dynamics} can be rewritten as

\[
\frac{\bi{x}_{k+1}-\bi{x}_{k}}{\Delta t}
=
\frac{v}{
\left\|
\nabla \bi{B}^{\mathsf T}_{k}
\left(\bi{B}_g-\bi{B}_{k}\right)
\right\|
}
\nabla \bi{B}^{\mathsf T}_{k}
\left(\bi{B}_g-\bi{B}_{k}\right).
\]

Taking the limit $\Delta t \to 0$, we obtain the continuous-time formulation
\[
\frac{\rmd \bi{x}}{\rmd t}
=
\eta(t)\,
\nabla \bi{B}^{\mathsf T}
\left(\bi{B}_g-\bi{B}\right),
\qquad \eta(t)>0,
\]
where
\[
\eta(t)
=
\frac{v}{
\left\|
\nabla \bi{B}^{\mathsf T}
\left(\bi{B}_g-\bi{B}\right)
\right\|
}.
\]

The positive scalar $\eta(t)$ controls the speed of motion without
changing the descent direction. For simplicity of the following
analysis, we set $\eta(t)=1$. The system dynamics then reduce to
\begin{equation}
\label{eq:potential_field_interpretation}
\frac{\rmd \bi{x}}{\rmd t}
=
\nabla \bi{B}^{\mathsf T}(\bi{B}_g-\bi{B})
=
-\frac{1}{2}\nabla V.
\end{equation}

Equation \eref{eq:potential_field_interpretation} therefore describes a gradient
flow, i.e., a continuous-time representation of gradient descent in
which the system evolves along the negative gradient of the objective
function $V$ toward its minimum. This property can be seen by evaluating
the variation of $V(\bi{x})$ along the system trajectory. Since $\bi{x}=\bi{x}(t)$, applying the chain rule gives
\[
\frac{\rmd V}{\rmd t}
=
\nabla V^{\mathsf T}\frac{\rmd \bi{x}}{\rmd t}.
\] Using \ref{eq:potential_field_interpretation}, it follows that
\[
\frac{\rmd V}{\rmd t}
= -\frac{1}{2}\nabla V^{\mathsf T}\nabla V =
-\frac{1}{2}\|\nabla V\|^2
\leq 0.
\]
Therefore, the objective function $V$ decreases along the system trajectory, showing that the gradient flow drives the system toward lower values of the objective function. This behavior drives the system toward an equilibrium state, which may correspond to the desired magnetic coordinates (the global minimum of $V$) or to another local minimum of $V$, where $\nabla V=0$. Hence, the system no longer moves according to \ref{eq:potential_field_interpretation}. This naturally provides stability at the equilibrium point. Since the system follows the negative gradient of $V$, any small displacement from the equilibrium causes the system to move toward lower values of $V$ and back toward the equilibrium. Under the assumed smoothness of the magnetic parameters, the stability of the gradient flow can be established using Lyapunov stability theory \parencite{absil_stable_2006}. However, stability alone does not guarantee reliable navigation, as the magnetic coordinates of the destination may not be unique. The potential $V$ may therefore contain multiple global minima or local minima that can trap the system. These problems arise as a consequence of the three main challenges described in the Introduction (section~\ref{sec:introduction}).

This motivates the approach proposed in this study, in which we aim to control the dynamics governed by \ref{eq:potential_field_interpretation}. We first note that, once the initial condition corresponding to the starting position is specified, the first differential equation, \ref{eq:potential_field_interpretation}, admits a unique solution by the Cauchy--Lipschitz theorem, provided that the magnetic parameters are sufficiently smooth \parencite{feng_note_2016}. The evolution of the path is governed by the magnetic environment, and the system follows its nominal dynamics, which do not necessarily guarantee convergence toward the desired destination characterized by the target magnetic field $\bi{B}_g$. This leads to the question of how to introduce a limited degree of control over this dynamical system while preserving its bio-inspired nature and its dependence on the magnetic environment. The locally measured parameters $\bi{B}$ and the estimated gradient $\nabla \bi{B}$ are dictated by the actual position of the system in the magnetic environment and cannot be changed. In contrast, $\bi{B}_g$ represents the magnetic coordinates of the desired destination and can be changed to represent another destination, thereby modifying the system trajectory. It is therefore of interest to determine a sequence of physical intermediate waypoints $\{\bi{w}\}_{i=1}^{n}$ within the magnetic environment that act as a form of control and guide the system toward the desired configuration, for example, by ensuring convergence to the actual destination and avoiding local minima. Since these waypoints belong to the magnetic environment, the bio-inspired nature of the navigation and its dependence on the magnetic environment are preserved. The system then navigates between successive waypoints by minimizing
\[
V_{\bi{w}_i} = \|\bi{B}_{\bi{w}_i}-\bi{B}\|^2
\]
according to the dynamics
\begin{equation}
\label{eq:waypoint_dynamic}
  \frac{\rmd \bi{x}}{\rmd t}
= 
\nabla \bi{B}^{\mathsf T}
\left(\bi{B}_{\bi{w}_i}-\bi{B}\right) = -\frac{1}{2}\nabla V_{\bi{w}_i},  
\end{equation}

where $\bi{B}_{\bi{w}_i}$ is the magnetic signature of the $i$th intermediate waypoint $\bi{w}_i$. The use of successive intermediate waypoints may initially appear artificial, but it is analogous to GPS-based navigation. For example, when navigating through a city, a GPS system does not necessarily guide the user directly toward the final destination. Instead, it follows a sequence of intermediate points to avoid obstacles and follow a feasible path constrained by the urban environment. A similar strategy is commonly used in everyday navigation, where intermediate landmarks facilitate movement from an origin to a final destination, e.g., ''turn at the pharmacy'' or ''walk until the park, then continue''. Similarly, the proposed approach uses intermediate magnetic waypoints to avoid local minima that can trap the system and to guide it toward the desired destination.

\subsection{Control and guidance of the gradient flow}
\label{sec:navigable_regions}
In this section, we define the regions where navigation paths can be constructed by guiding the gradient flow through intermediate destinations. These regions are defined a priori as areas where theoretical guarantees can be established for the existence of a sequence of intermediate waypoints that controls the system dynamics. In BIMN, we use magnetic parameters as local coordinates to navigate over the two-dimensional surface of the Earth, which is known in the literature as the bi-coordinate map \parencite{benhamou_bicoordinate_2003, bostrom_where_2012}. For these parameters to provide two independent spatial coordinates, the gradient matrix
\[
\nabla \bi{B} =
\left(
\begin{array}{cc}
\displaystyle \frac{\partial I}{\partial x} &
\displaystyle \frac{\partial I}{\partial y} \\[4pt]
\displaystyle \frac{\partial F}{\partial x} &
\displaystyle \frac{\partial F}{\partial y}
\end{array}
\right)
\]
must have full rank, i.e., $\mathop{\rm rank}(\nabla \bi{B}) = 2.$ The gradient matrix $\nabla \bi{B}$ has maximum rank if and only if its determinant $\det(\nabla \bi{B})$ is non-zero. We show that, within such regions, it is possible to navigate toward any destination through a sequence of intermediate waypoints, provided that the path does not cross frontiers where $\det(\nabla \bi{B})=0$. In other words, these frontiers correspond to a degeneracy of the maximum rank of the gradient, separating regions characterized by opposite signs of the determinant. Since the determinant is continuous, it must be zero between these regions, according to the intermediate value theorem. Based on this, we define a navigable region as a connected area characterized by a constant sign of $\det(\nabla \bi{B})$, in which any two positions can be connected by a path without crossing the frontiers. Such region is suitable for navigation for two main reasons. First, the gradient $\nabla \bi{B}$ is invertible. When $\det(\nabla\bi{B})\neq 0$, the inverse function theorem ensures that the mapping from the spatial position $\bi{x}=(x,y)$ to the magnetic parameters $\bi{B}=(I,F)$ is locally one-to-one. In other words, around any point in the defined region, each position is locally characterized by a unique set of magnetic parameters. Thus, two nearby positions cannot have the same magnetic parameters.

The second reason is that the objective function
\[
\|\bi{B}_{\bi{w}}-\bi{B}(\bi{x})\|^2
\]
is locally strongly convex around any point $\bi{w}$ within the defined region. Intuitively, this means that $\bi{w}$ acts as a local attractor, pulling the gradient flow toward it. To see this, we use a first-order Taylor expansion of $\bi{B}(\bi{x})$ around $\bi{w}$:
\[
\bi{B}(\bi{x})
\approx
\bi{B}_{\bi{w}}
+
\nabla \bi{B}_{\bi{w}}(\bi{x}-\bi{w}).
\]

Therefore,
\[
\|\bi{B}_{\bi{w}}-\bi{B}(\bi{x})\|^2
\approx
(\bi{x}-\bi{w})^{\mathsf T}
\nabla\bi{B}_{\bi{w}}^{\mathsf T}
\nabla\bi{B}_{\bi{w}}
(\bi{x}-\bi{w}).
\]

Let $\bi{\widetilde{w}}=\bi{x}-\bi{w}$. The quadratic approximation then becomes
\[
\bi{\widetilde{w}}^{\mathsf T}
\nabla\bi{B}_{\bi{w}}^{\mathsf T}
\nabla\bi{B}_{\bi{w}}
\bi{\widetilde{w}}.
\]

Since $\nabla\bi{B}_{\bi{w}}$ has maximum rank,
$\nabla\bi{B}_{\bi{w}}^{\mathsf T}\nabla\bi{B}_{\bi{w}}$
is symmetric positive definite. Therefore, by the spectral theorem, it is diagonalizable and all its eigenvalues are strictly positive.  If $\lambda_1$ and $\lambda_2$ denote these eigenvalues, then
\[
\bi{\widetilde{w}}^{\mathsf T}
\nabla\bi{B}_{\bi{w}}^{\mathsf T}
\nabla\bi{B}_{\bi{w}}
\bi{\widetilde{w}}
\geq
\min(\lambda_1,\lambda_2)\|\bi{\widetilde{w}}\|^2.
\] This shows that the objective function is locally strongly convex around $\bi{w}$, which guarantees convergence of the system to $\bi{w}$.

These two arguments allow us to guide the system toward a desired destination using a sequence of intermediate waypoints. Although the above guarantees are local, the waypoints can be chosen sufficiently close to one another so that each waypoint lies within the local region of attraction of the next one. In this way, a navigation path can be constructed by successively guiding the system from one waypoint to the next until the final destination is reached. This provides, at least locally, a theoretical basis for the existence of a suitable sequence of control waypoints. In contrast, these guarantees are lost when approaching the frontiers where $\det(\nabla\bi{B})=0$. However, this does not mean that these frontiers cannot be crossed; crossing them can simply be difficult. At these frontiers, the gradient $\nabla\bi{B}$ does not have full rank and therefore has one or more null directions. As a result, the dynamics $\nabla\bi{B}^{\mathsf T}(\bi{B}_{\bi{w}}-\bi{B})$ can vanish along the frontier if $\bi{B}_{\bi{w}}$ is not chosen appropriately to avoid these null directions. In practice, measurement noise can make this situation even more difficult by causing $(\bi{B}_{\bi{w}}-\bi{B})$ to point along, or close to, a null direction of the gradient, resulting in little or no motion. Even with a suitable choice of $\bi{B}_{\bi{w}}$, crossing may still be impossible. For example, if $\nabla\bi{B}$ has rank one, only one direction of motion remains at the frontier. If this direction is transverse to the frontier and points across it, crossing is possible, although with the risks mentioned above. If the direction is tangent to the frontier, the system can only move along the frontier and cannot cross it using the gradient descent dynamics alone. This is why we focus on navigation within the defined regions, where the gradient has full rank and provides guarantees, rather than on crossing the frontiers that separate these regions, where these guarantees are lost and the risk is higher. In particular, for the gradient descent dynamics in equation~\ref{eq:waypoint_dynamic}, these frontiers contain local minima of the objective function $V_{{w}}$, where $\nabla V_{\bi{w}}=0$, at which the system may become trapped.

\subsection{Path optimization}
\label{sec:path_optimization}
The use of intermediate waypoints is fundamental for guiding the gradient flow and ensuring that the system converges to the destination. However, the path between two successive waypoints may not be efficient in terms of distance traveled. The system follows the direction of steepest decrease of $V$, which does not necessarily coincide with the shortest direction toward the target waypoint. Two main factors can contribute to this. First, the terms $(I_g-I)^2$ and $(F_g-F)^2$ can have different scales and therefore contribute unequally to $V$ and its gradient. If one contribution dominates, the descent direction is primarily determined by the corresponding magnetic parameter. Second, the spatial gradients of $I$ and $F$ can be strongly direction dependent. For example, if $|\partial F/\partial x| \gg |\partial F/\partial y|$, variations in $F$ contribute much more strongly to the gradient of $V$ along the $x$-direction than along the $y$-direction. A similar anisotropy may occur for $I$. Consequently, the descent direction can be preferentially oriented along directions in which the magnetic parameters vary most rapidly, rather than toward the target waypoint. This can further produce longer or more indirect trajectories. To mitigate this
effect, we introduce two positive adjustable parameters for each path segment connecting
two consecutive waypoints. These parameters regulate the relative contributions
of the magnetic quantities and their spatial variations, thereby improving the
resulting trajectory. Let
$
\bi{B}_{\bi{w}} = (I_\bi{w},F_\bi{w})
$
denote the magnetic parameters associated with the destination of a given path segment. We then define the weighted objective function as
\[
V_{\alpha_\bi{w}}
= \alpha_\bi{w} (I_\bi{w}-I)^2 + (F_\bi{w}-F)^2,
\]
where $\alpha_\bi{w}$ controls the relative contribution of the inclination term with respect to the intensity term. The corresponding modified flow is defined as
\[
\frac{\rmd \bi{x}}{\rmd t}
=
\left(
\begin{array}{cc}
\displaystyle \frac{\partial I}{\partial x}
&
\displaystyle \frac{\partial F}{\partial x}
\\[4pt]
\displaystyle \beta_\bi{w}\frac{\partial I}{\partial y}
&
\displaystyle \beta_\bi{w}\frac{\partial F}{\partial y}
\end{array}
\right)
\left(
\begin{array}{c}
\displaystyle \alpha_\bi{w}(I_\bi{w}-I)
\\[4pt]
\displaystyle F_\bi{w}-F
\end{array}
\right),
\]
where $\beta_\bi{w}$ adjusts the relative contribution of the spatial variations along
the $y$-direction with respect to those along the $x$-direction. These two introduced parameters do not change the navigable regions defined in
section~\ref{sec:navigable_regions}. Indeed, the modified gradient flow can be written as
\[
\frac{\rmd \bi{x}}{\rmd t}
=
\bi{J}_\bi{w}
\left(
\begin{array}{c}
I_\bi{w}-I\\
F_\bi{w}-F
\end{array}
\right),
\]
where
\[
\bi{J}_\bi{w}
=
\left(
\begin{array}{cc}
\displaystyle
\alpha_\bi{w}\frac{\partial I}{\partial x}
&
\displaystyle
\frac{\partial F}{\partial x}
\\[4pt]
\displaystyle
\alpha_\bi{w}\beta_\bi{w}\frac{\partial I}{\partial y}
&
\displaystyle
\beta_\bi{w}\frac{\partial F}{\partial y}
\end{array}
\right).
\]
The determinant of the modified gradient matrix $\bi{J}_\bi{w}$ is therefore given by
\[
\begin{array}{rcl}
\det(\bi{J}_\bi{w})
& = &
\displaystyle
\alpha_\bi{w}\beta_\bi{w}
\left(
\frac{\partial I}{\partial x}
\frac{\partial F}{\partial y}
-
\frac{\partial F}{\partial x}
\frac{\partial I}{\partial y}
\right)
\\[6pt]
& = &
\displaystyle
\alpha_\bi{w}\beta_\bi{w} \det(\nabla \bi{B}).
\end{array}
\]
Consequently, provided that $\alpha_\bi{w}>0$ and $\beta_\bi{w}>0$, the determinant
vanishes at exactly the same locations as in the original formulation of the gradient flow in \ref{eq:potential_field_interpretation}. Therefore, the introduction of $\alpha_\bi{w}$ and $\beta_\bi{w}$ does not alter the frontiers or topology of the navigable regions; it only modifies the structure of the gradient flow within these regions. 

The modified flow remains a descent flow for $V_{\alpha_\bi{w}}$. To show this,
consider a local direction of motion $\bi{d}$ and a sufficiently small step
$\epsilon>0$. A first-order Taylor expansion gives
\[
V_{\alpha_\bi{w}}(\bi{x}+\epsilon\bi{d})
=
V_{\alpha_\bi{w}}(\bi{x})
+
\epsilon\,\nabla V_{\alpha_\bi{w}}^{\mathsf T}\bi{d}.
\]
Thus, $V_{\alpha_\bi{w}}$ decreases locally whenever
$\nabla V_{\alpha_\bi{w}}^{\mathsf T}\bi{d}<0$.

For the weighted objective function
\[
V_{\alpha_\bi{w}}
=
\alpha_\bi{w}(I_\bi{w}-I)^2+(F_\bi{w}-F)^2,
\]
the gradient is
\[
\nabla V_{\alpha_\bi{w}}
=
-2
\left(
\begin{array}{c}
\displaystyle
\alpha_\bi{w}(I_\bi{w}-I)\frac{\partial I}{\partial x}
+
(F_\bi{w}-F)\frac{\partial F}{\partial x}
\\[6pt]
\displaystyle
\alpha_\bi{w}(I_\bi{w}-I)\frac{\partial I}{\partial y}
+
(F_\bi{w}-F)\frac{\partial F}{\partial y}
\end{array}
\right).
\]
On the other hand, the modified flow is given by
\[
\frac{\rmd\bi{x}}{\rmd t}
=
\left(
\begin{array}{c}
\displaystyle
\alpha_\bi{w}(I_\bi{w}-I)\frac{\partial I}{\partial x}
+
(F_\bi{w}-F)\frac{\partial F}{\partial x}
\\[6pt]
\displaystyle
\beta_\bi{w}
\left[
\alpha_\bi{w}(I_\bi{w}-I)\frac{\partial I}{\partial y}
+
(F_\bi{w}-F)\frac{\partial F}{\partial y}
\right]
\end{array}
\right).
\]
Therefore,
\[
\nabla V_{\alpha_\bi{w}}^{\mathsf T}
\frac{\rmd\bi{x}}{\rmd t}
=
-\frac{1}{2}
\left[
\left(\frac{\partial V_{\alpha_\bi{w}}}{\partial x}\right)^2
+
\beta_\bi{w}
\left(\frac{\partial V_{\alpha_\bi{w}}}{\partial y}\right)^2
\right].
\]
Hence, for $\beta_\bi{w}>0$,
\[
\frac{\rmd V_{\alpha_\bi{w}}}{\rmd t}\leq 0,
\]
with equality only when $\nabla V_{\alpha_\bi{w}}=\bi{0}$. Therefore, the
modified flow preserves the descent property of the original formulation. The parameter $\alpha_\bi{w}$ adjusts the relative contribution of
the inclination error with respect to the intensity error in the objective
function, while $\beta_\bi{w}$ adjusts the relative contribution of the
spatial variations along the $y$-direction with respect to those along the
$x$-direction. Provided that $\alpha_\bi{w}>0$ and $\beta_\bi{w}>0$, these
modifications preserve the decreasing behavior of $V_{\alpha_\bi{w}}$ while
allowing the geometry of the resulting trajectory to be adjusted within
the navigable region.

\paragraph{}
In summary, the proposed BIMN framework, based on guidance and control of the gradient flow, is reliable within the defined navigable regions. We have shown that, within these regions, a sequence of intermediate waypoints can always be selected to reach the destination. For each path segment, two additional parameters can be tuned to optimize the shape of the overall trajectory. From a bio-inspired perspective, these intermediate waypoints may be viewed as preferred locations acquired by animals through evolution and learning. In practice, however, determining the intermediate waypoints and their associated path optimization parameters online would require extensive exploration of the environment. For an autonomous system, such exploration is constrained by the available energy and, more importantly, by safety requirements.

A more practical solution is therefore to select the intermediate waypoints from
an available magnetic map, such as the IGRF
\parencite{beggan_international_2026}, and to optimize their associated parameters
offline. The resulting navigation plan is then reduced to a small set of
intermediate waypoints and parameters that can be stored onboard. Thus, the magnetic map does not need to be stored onboard or accessed during navigation.
% \FloatBarrier
\clearpage
\section{Results}
\label{sec:results}
In this section, we perform simulations using the magnetic inclination and field intensity obtained from the International Geomagnetic Reference Field (IGRF) \parencite{beggan_international_2026}. The magnetic parameters are evaluated at an altitude of 10 km, representative of aircraft navigation, and at the reference date 01/01/2026. The Earth's main magnetic field evolves slowly with time. According to the IGRF, secular variation is generally on the order of tens to about \(150\,\mathrm{nT/year}\), compared with a total field intensity of approximately \(22\)–\(70\,\mathrm{\mu T}\). Therefore, a one-year change in the reference date produces only a small relative change in the field intensity \parencite{beggan_international_2026}.

\subsection{Control and guidance of the gradient flow}
\begin{figure*}[t]
\centering
\fbox{%
\begin{minipage}{\dimexpr\textwidth-2\fboxsep-2\fboxrule\relax}
\centering

\includegraphics[width=\linewidth]
{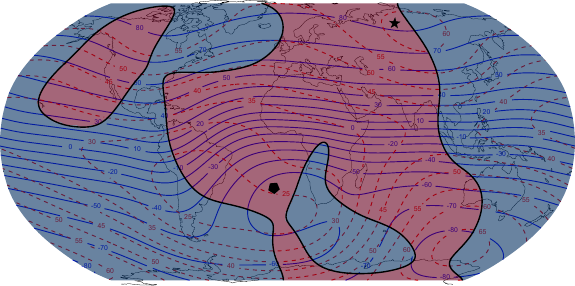}

\caption{Navigable regions, where the continuous black lines define the frontiers $\det(\nabla \bi{B})=0$, the gray-blue areas indicate regions with $\det(\nabla \bi{B})>0$, and the gray-red areas indicate regions with $\det(\nabla \bi{B})<0$. These regions (colored gray-blue and gray-red) are separated by frontiers where the gradient exhibits a degeneracy. There are four separated areas suitable for navigation: two gray-blue and two gray-red. The starting position is indicated by a star, while the final destination is indicated by a pentagon. Both lie within the largest gray-red region.}
\label{fig:navigable_zones}

\end{minipage}%
}
\end{figure*}

\begin{figure*}[t]
\centering
\fbox{%
\begin{minipage}{\dimexpr\textwidth-2\fboxsep-2\fboxrule\relax}
\centering

% First row: (a) and (b)
\begin{minipage}[b]{0.48\linewidth}
    \centering
    \includegraphics[width=\linewidth]{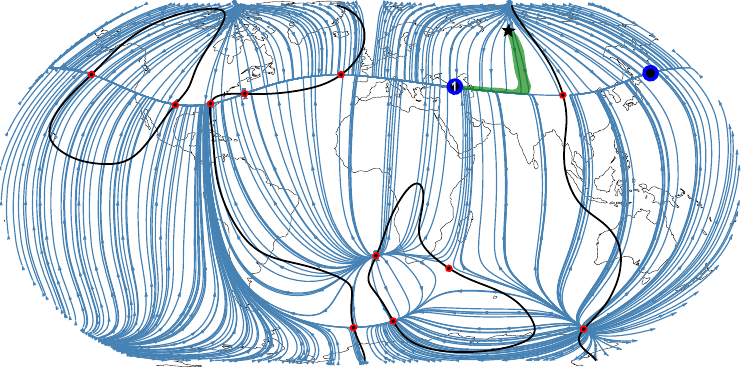}\\[2pt]
    \textbf{(a)}
\end{minipage}
\hfill
\begin{minipage}[b]{0.48\linewidth}
    \centering
    \includegraphics[width=\linewidth]{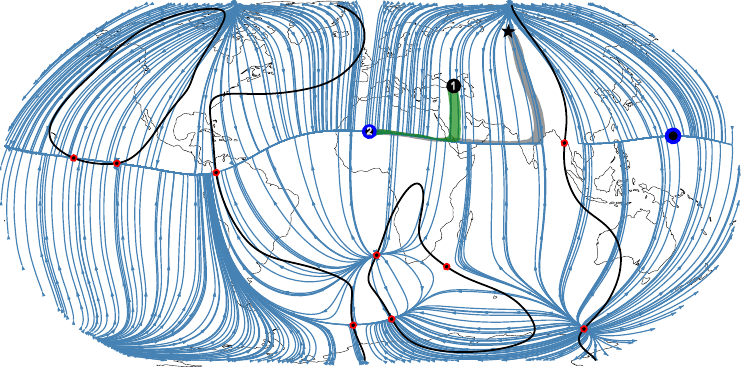}\\[2pt]
    \textbf{(b)}
\end{minipage}

\vspace{1em}

% Second row: (c) and (d)
\begin{minipage}[b]{0.48\linewidth}
    \centering
    \includegraphics[width=\linewidth]{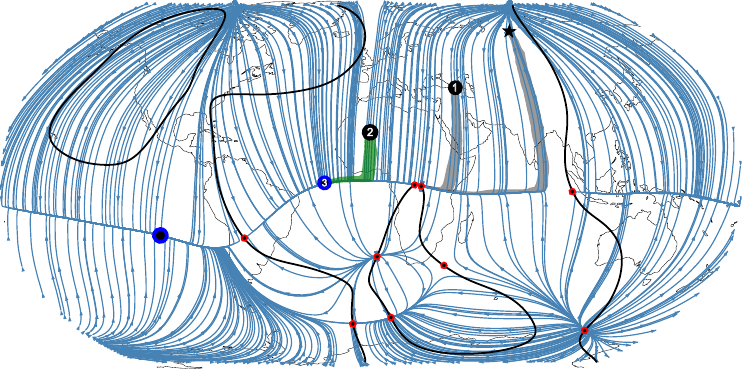}\\[2pt]
    \textbf{(c)}
\end{minipage}
\hfill
\begin{minipage}[b]{0.48\linewidth}
    \centering
    \includegraphics[width=\linewidth]{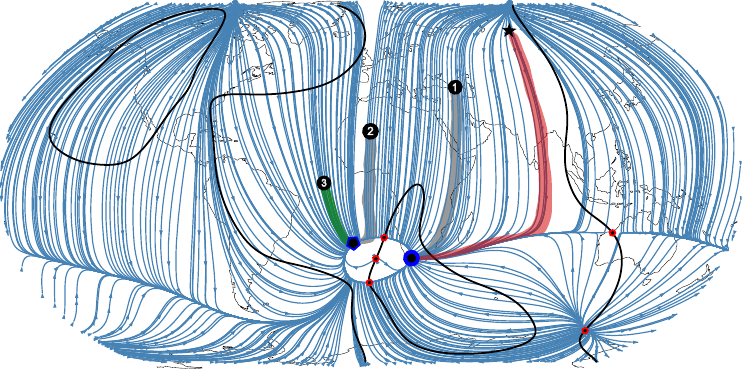}\\[2pt]
    \textbf{(d)}
\end{minipage}

\caption{Example of controlling the flow to guide the system toward the destination. The initial path from the starting position, controlled by the magnetic parameters of the destination $\bi{B}_g$, converges to another location, as shown in (d) by the red-gray path. We therefore guide the flow sequentially through three intermediate waypoints. From the starting position, the flow is first guided to $\bi{w}_1$ using its magnetic coordinates $\bi{B}_{\bi{w}_1}$, as shown by the green-gray path in (a). The flow is then guided from $\bi{w}_1$ to $\bi{w}_2$ using $\bi{B}_{\bi{w}_2}$ in (b), from $\bi{w}_2$ to $\bi{w}_3$ using $\bi{B}_{\bi{w}_3}$ in (c), and finally from $\bi{w}_3$ to the destination using $\bi{B}_g$ in (d). The two blue points in each subfigure share the same magnetic coordinates and correspond to global minima. The red points on the frontiers are critical points, including local minima, local maxima, and saddle points, visually showing that crossing these frontiers is challenging for the gradient descent navigation mechanism. Navigation from $\bi{w}_1$ to the final destination is not possible, as the flow converges to the fake destination. Therefore, we introduce $\bi{w}_2$. Navigation from $\bi{w}_2$ to the final destination is possible, but we introduce $\bi{w}_3$ to steer the trajectory away from the frontiers and avoid the critical points. We can see that direct navigation from either the starting point or $\bi{w}_1$ to $\bi{w}_3$ is not possible, as the corresponding flows, shown in gray in (c), converge to a critical point. Similarly, the flow from $\bi{w}_1$ to the final destination converges to the fake destination, as shown in gray in (d). This highlights the importance of introducing successive intermediate waypoints to guide the flow toward the real final destination and avoid these difficulties.}
\label{fig:control_and_guidance_of_gradient_flow}

\end{minipage}%
}
\end{figure*}

Figure~\ref{fig:navigable_zones} shows the topology and extent of the navigable regions, which cover a large portion of the Earth's surface. These regions are obtained by numerically evaluating the gradient $\nabla\bi{B}$ using the IGRF and computing its determinant. The phase portrait of the gradient flow is obtained numerically by first evaluating the vector field associated with \ref{eq:potential_field_interpretation} over a spatial grid and then integrating the corresponding differential equation from a set of initial conditions using the Runge--Kutta--Fehlberg method of order 5(4) (RK45). We then show an example of controlling the gradient flow between two locations. The starting position has GPS coordinates (longitude, latitude) of $\bi{x}_s = (89^\circ, 66^\circ)$, and the final destination is located at $\bi{x}_g = (-8^\circ, -26^\circ)$, with magnetic coordinates $\bi{B}_g = (-64^\circ, 25~\mu\mathrm{T})$. Both positions are shown in figure~\ref{fig:navigable_zones}. Direct navigation between these two points is not feasible, as the gradient flow from the starting position, controlled by $\bi{B}_g$ without intermediate destinations, converges to another location that has the same magnetic coordinates (see figure~\ref{fig:control_and_guidance_of_gradient_flow}(d), shown in red-grey). We used three intermediate destinations to guide the gradient flow toward the final destination. Figure~\ref{fig:control_and_guidance_of_gradient_flow} shows the gradient flow computed without optimization, by setting $\alpha=1$ and $\beta=1$, as defined in section~\ref{sec:path_optimization}, for all segments of the navigation path. The intermediate waypoints $\bi{w}_1$, $\bi{w}_2$, and $\bi{w}_3$ have GPS coordinates and magnetic coordinates, respectively,
$\bi{x}_{\bi{w}_1} = (44^\circ, 40^\circ)$, $\bi{x}_{\bi{w}_2} = (0^\circ, 21.93^\circ)$, $\bi{x}_{\bi{w}_3} = (-22.48^\circ, 0^\circ)$,
and
$\bi{B}_{\bi{w}_1} = (59^\circ, 49~\mu\mathrm{T})$, $\bi{B}_{\bi{w}_2} = (24.41^\circ, 36.91~\mu\mathrm{T})$, $\bi{B}_{\bi{w}_3} = (-26.84^\circ, 28.88~\mu\mathrm{T})$.

\paragraph{}
The gradient flow guidance simulation uses three waypoints as control nodes to shape the navigation path. Additional control nodes can be introduced to further adjust the path and improve navigation robustness. This robustness is critical for compensating for heading errors encountered in real-world navigation. As long as the system remains within the region of attraction of the target waypoint, an onboard gradient estimate could be updated online based on the mismatch between the predicted and measured magnetic parameters caused by the heading error, allowing the estimated gradient to accommodate heading bias and maintain convergence toward the target. Visually, this can be seen in the figure~\ref{fig:control_and_guidance_of_gradient_flow}, where the nominal flow trajectory is surrounded by neighboring trajectories that also converge toward the target waypoint. Thus, a heading error may shift the system from the nominal trajectory to a nearby trajectory. As long as this trajectory remains within the region of attraction, the system still converges toward the target.

% parlle reééllment du probelem de descent de gradient 
% now you can mentione the descrete version....by refering back 
Figure~\ref{fig:optimized_not_optimized_path} shows the navigation path without optimization and the optimized path obtained by manually tuning the $\alpha$ and $\beta$ parameters associated with each path segment, while keeping the three previous waypoints fixed (Table~\ref{tab:optimization_parameters}). These two trajectories are obtained using the discrete-time dynamics in \ref{eq:discrete_dynamics}, with a constant velocity of
$v = 100~\mathrm{km\,h^{-1}}$ and a time step of $\Delta t = 0.05~\mathrm{h}$. Thus, the system moves $5~\mathrm{km}$ between two consecutive updates. At each update, a new measurement is taken and the gradient is accessed. A waypoint is considered reached when the magnetic mismatch satisfies
\[
\left\|\bi{B}_g-\bi{B}\right\| \leq 0.2.
\]
Once this condition is satisfied, the next waypoint is selected as the new destination. This process is repeated until the final destination is reached. In this example, we constructed a single magnetic route. This could be extended to form a network of interconnected magnetic routes sharing common waypoints, thereby covering a larger region and providing an alternative to satellite navigation.

\begin{table}[ht]
\centering

\setlength{\tabcolsep}{8pt}
\renewcommand{\arraystretch}{1.5}

\begin{tabular}{|c|c|c|c|c|}
\cline{2-5}
\multicolumn{1}{c|}{}
 & Start--$\bi{w}_1$
 & $\bi{w}_1$--$\bi{w}_2$
 & $\bi{w}_2$--$\bi{w}_3$
 & $\bi{w}_3$--Dest. \\
\hline
$\alpha$ & 0.25 & 0.20 & 0.10 & 1.00 \\
\hline
$\beta$  & 1.00 & 0.10 & 0.10 & 1.00 \\
\hline
\end{tabular}

\caption{Manually selected optimization parameters for each navigation segment.}
\label{tab:optimization_parameters}
\end{table}

Note that in figure~\ref{fig:control_and_guidance_of_gradient_flow}, the single intermediate waypoint $\bi{w}_2$ is sufficient to connect the starting position to the final destination. The flow from the starting position, controlled by $\bi{B}_{\bi{w}_2}$, converges to $\bi{w}_2$, as shown in gray in figure~\ref{fig:control_and_guidance_of_gradient_flow}(b). From $\bi{w}_2$, the flow controlled by the destination parameters $\bi{B}_g$ converges to the final destination, as shown in gray in figure~\ref{fig:control_and_guidance_of_gradient_flow}(d). However, introducing additional intermediate waypoints can provide better control over the path and improve the efficiency of the optimization parameters. Shorter path segments allow the local parameterization of $\alpha$ and $\beta$ to better capture and compensate for the complexity of the magnetic environment.

\subsection{Implications of observed sea turtle migration paths}
\begin{figure*}[t]
\centering
\fbox{%
\begin{minipage}{\dimexpr\textwidth-2\fboxsep-2\fboxrule\relax}
\centering

\includegraphics[width=\linewidth]
{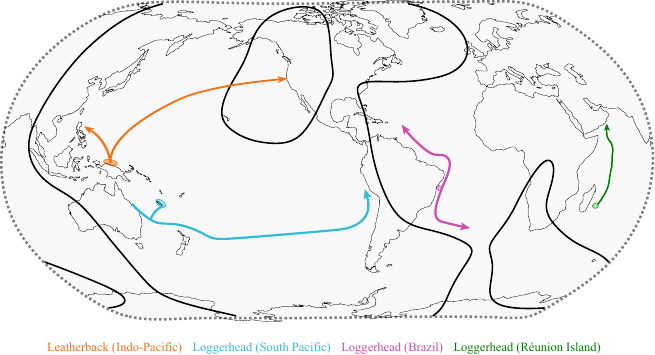}

\caption{Schematic representation of iconic animal migration routes, showing the general migration patterns rather than precise trajectories, as reported by \parencite{putman_marine_2018} for the Indo-Pacific leatherback, South Pacific loggerhead, and Brazilian loggerhead turtles, and by \parencite{laforge_uncovering_2023} for the Réunion Island loggerhead turtle.}
\label{fig:iconic_turtles_migrations_paths}

\end{minipage}%
}
\end{figure*}

Figure~\ref{fig:iconic_turtles_migrations_paths} shows four iconic migration paths of leatherback and loggerhead turtles, together with the frontiers of the navigable regions. Three loggerhead turtle migration paths, corresponding to the South Pacific, Brazil, and Réunion Island, lie within these navigable zones. This suggests that it is plausible that these animals may rely on the Earth's magnetic field for navigation. As demonstrated by the proposed BIMN framework, intermediate waypoints can be introduced, when necessary, to guide the path toward the final destination. In natural turtle migration, such intermediate waypoints may correspond to feeding or resting sites learned through their evolutionary history. From the perspective of the proposed framework, preserving such sites may therefore be important for maintaining established migration routes and, ultimately, for the survival of these species. Given the relatively large size of these species, GPS tracking devices can be deployed to monitor their migration trajectories with minimal disturbance to the animals. The resulting data could be used to identify precise intermediate waypoints, under the hypothesis that the Earth's magnetic field serves as a primary navigation cue. Combining the proposed framework with existing models that account for ocean currents, water temperature, and the availability of feeding sites \parencite{luschi_long-distance_2013, laforge_uncovering_2023} could provide a more comprehensive understanding of turtle migration mechanisms. Such an approach could help protect these species from increasing disturbances caused by human activities along their migration routes, for example, by predicting migration paths and restricting fishing activities within identified migration corridors \parencite{baudouin_identification_2015, iverson_migration_2020}.

\paragraph{}
The Indo-Pacific migration path of the leatherback turtle does not remain entirely within a navigable region, as it crosses a frontier characterized by gradient degeneracy. As discussed in section~\ref{sec:navigable_regions}, crossing such frontiers is not necessarily impossible, but it can be challenging when relying uniquely on magnetic navigation. However, sea turtles are multimodal navigators and may switch to other navigation cues when magnetic navigation becomes challenging. This limitation could be considered together with the navigation difficulties already reported in the literature for leatherback turtles, potentially providing a complementary explanation for observed anomalies in their migration paths \parencite{sale_navigational_2009, benson_large-scale_2011}.

\section{Discussion}
\begin{figure*}[t]
\centering
\fbox{%
\begin{minipage}{0.82\textwidth}
\centering

\includegraphics[width=\linewidth]
{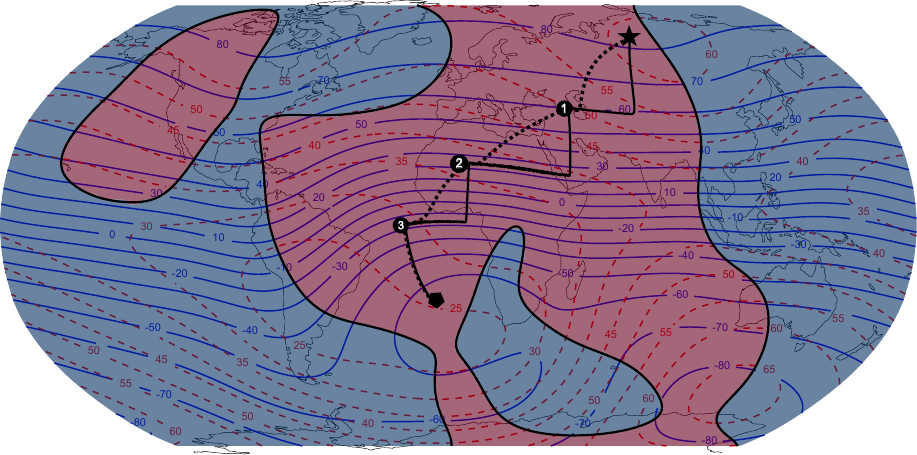}

\medskip

\includegraphics[width=\linewidth]
{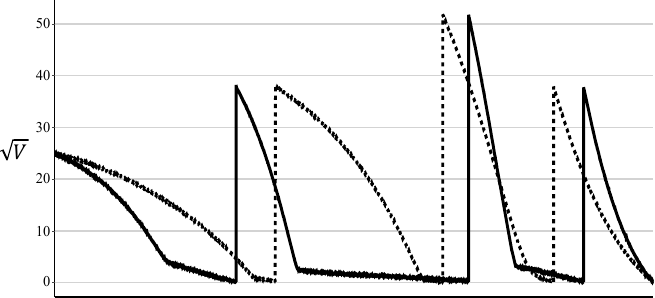}

\caption{The optimized path obtained using the parameters in table~\ref{tab:optimization_parameters} is shown by the dashed black line, while the original path with $\alpha=\beta=1$ is shown by the solid black line. The optimized path, obtained by adjusting $\alpha$ and $\beta$ for each path segment between two successive waypoints, provides a better trajectory than the original path. The evolution of the objective function $V$ for each path segment is also shown. For both paths, $V$ decreases within each segment and switches to the objective associated with the next target waypoint when the current waypoint is reached, resulting in the peaks observed in the figure. The original path initially exhibits a faster decrease in $V$, as it follows the direction of steepest descent. In contrast, the optimized path decreases $V$ more gradually, as the parameters $\alpha$ and $\beta$ modify the local flow to avoid unnecessarily long paths while maintaining a decreasing direction toward the target. Thus, a faster decrease in $V$ does not necessarily correspond to a better navigation path. The vertical axis represents $\sqrt{V}$, which is treated as a dimensionless objective function. The horizontal axis represents the elapsed navigation time, with the system moving at a constant velocity of $v = 100~\mathrm{km\,h^{-1}}$ with a sampling period $\Delta t = 0.05~\mathrm{h}$. Since the main interest is the evolution of the objective function, the time values are not explicitly shown.}
\label{fig:optimized_not_optimized_path}

\end{minipage}%
}
\end{figure*}

The planned magnetic navigation route shown in Fig.~\ref{fig:optimized_not_optimized_path} is selected from the magnetic map assuming perfect knowledge of the gradients of the magnetic parameters. In practice, these gradients must be estimated, and estimation errors may cause the actual navigation path to deviate from the planned route. Therefore, Monte Carlo simulations with measurement noise, heading drifts, and magnetic perturbations can be used to represent more realistic conditions and to quantify the variability of the navigation performance under uncertainty. The navigation method in \parencite{herma_bio-inspired_2026}, based on gradient estimation and gradient descent, is then used to test the robustness of the selected navigation path. In addition, the spatial resolution of the magnetic map may affect the accuracy of the planned route. Filtering and adaptive motion strategies can help reduce this effect. For example, the system velocity or sampling period can be adjusted according to local magnetic field variations, which changes the step size $v\Delta t$ in \ref{eq:discrete_dynamics}. If the magnetic map has a low spatial resolution and mainly represents large-scale variations, as in the IGRF, a larger step size can be used to reduce the influence of small local variations. This can be combined with online smoothing or filtering techniques to reduce the mismatch between the measured magnetic coordinates and the magnetic map used for planning the navigation route.

\paragraph{}
The selection of intermediate waypoints should account for both the spatial resolution of the magnetic map and the information it contains. Since some regions may exhibit finer spatial variations than others, the sampling period, and therefore the step size, can be adapted along the path between successive waypoints to account for such variations. This is not the case in the present study, in which a fixed step size is used to navigate through all intermediate waypoints along the planned route shown in \ref{fig:optimized_not_optimized_path}.

\paragraph{}
The planned path should also account for the information available in the magnetic map and the reliability of the estimation and filtering methods. In particular, it may be advantageous to favor regions with strong gradient, as these provide more informative measurements and can improve gradient estimation. On the other hand, regions where the gradient estimation or filtering is less reliable may lead to larger navigation errors and should therefore be avoided when possible. For example, if the sensor used for navigation has a lower resolution than the magnetic map used for planning, it is preferable to pass through regions with strong magnetic gradients. This allows the sensor to measure magnetic changes as the system moves, depending on the sampling period and the system's velocity.

\paragraph{}
The stopping criterion, which determines whether the system has reached the destination in the magnetic coordinate space, directly affects the achievable navigation accuracy. It also depends on the magnetic map used for path planning, since the map resolution introduces uncertainty in the estimated distance to the destination. The intermediate waypoints $\bi{w}_i$, optimization parameters $\alpha_i$ and $\beta_i$, and step size $(v\Delta t)_i$ can also be refined using optimal control, as they can be treated as control variables that influence the planned route and the accuracy of reaching the target.

\paragraph{}
The navigation frame is not discussed in this work, but it is important in practice because the heading is calculated relative to a reference direction. Magnetic north can be used as a reference, but its measured direction using a compass varies from one location to another. Similarly, a gyroscope measures angular velocity, which can be integrated to estimate the heading. However, errors accumulate over time, leading to drift. Other orientation technologies also have their own limitations and sources of error. The advantage of BIMN is its apparent robustness to changes or errors in the navigation frame heading \parencite{qi_geographic_2023}. This can be understood by considering that navigation takes place in a measurement space defined by magnetic coordinates, where the navigation process involves minimization. This minimization can compensate for heading errors in the estimated gradient \parencite{herma_bio-inspired_2026}.

\paragraph{}
The proposed BIMN framework operates in regions where the gradient matrix has full rank, as shown in \ref{fig:navigable_zones}. Crossing the frontiers of these regions can therefore be challenging. To address this limitation, the framework can be extended to include gravity measurements as additional parameters. This may increase the rank of the gradient matrix and increase the size of the navigable regions. Another, simpler approach is to temporarily switch to compass-based navigation when crossing these frontiers. Once the system reaches the other side, where the magnetic gradient again has full rank, it can return to BIMN using an intermediate waypoints defined in the new region. The main challenge during this transition is directional drift, as the compass heading may be affected by local magnetic anomalies and spatial variations in magnetic declination, since the Earth's magnetic field is not a perfect dipole. However, once the system enters the new region, the magnetic gradient can be re-estimated, and gradient descent toward the next waypoint can correct the accumulated directional error and guide the system back toward its desired goal.

\section{Conclusion}

In conclusion, this work demonstrates that gradient flows of magnetic inclination and intensity can serve as a GPS-free guidance mechanism. We extend the gradient descent method developed in \parencite{herma_bio-inspired_2026} and formulate BIMN as a gradient flow control and guidance problem. Waypoints are introduced to shape the gradient flow and guide the system along the desired path within the navigable region, defined by the maximum gradient rank. The proposed BIMN framework can enhance autonomous navigation in engineering applications by providing an alternative to vulnerable satellite navigation systems. It may also serve as a complementary approach for studying animal migration trajectories, as many animals are known to use the Earth’s magnetic field as a natural navigation cue.
\paragraph{}
Future work will focus on automated planning and optimization of magnetic routes using optimal control to determine the placement of magnetic waypoints and the associated path parameters. This will enable the desired navigation path to be followed and the target to be reached without manual tuning of these control variables. This formulation could also incorporate geodesic trajectories to obtain more direct paths toward the desired configuration.

The magnetic environment considered in this work should also be extended to account for temporal variations, such as those associated with solar activity, as well as crustal magnetic fields arising from the Earth’s crust. These variations may introduce disturbances that affect magnetic navigation. Online filtering methods could therefore be incorporated to reduce their effects and improve the reliability of the measured magnetic parameters. Similarly, animals that rely on magnetoreception may employ biological filtering mechanisms to attenuate magnetic disturbances and extract reliable cues for navigation.

The analysis of animal migration paths should also be extended using publicly available GPS data to identify waypoints associated with feeding sites, using the IGRF model evaluated at the altitude and time relevant to the specific type of migration. This would allow a more precise analysis of the migration trajectories. The long-term evolution of the Earth’s magnetic field could also be investigated to assess its possible impact on the evolution of migration paths over several years.

Finally, the simulations presented here assume perfect access to the gradients of the magnetic parameters provided by the IGRF. Future studies should consider a more realistic magnetic environment, including measurement noise and disturbances in the estimated gradient direction, such as those caused by ocean currents. The gradient could be estimated directly from magnetic measurements, as in the method developed in \parencite{herma_bio-inspired_2026}. The effects of solar activity and other magnetic disturbances on the generated trajectories could also be included. This would provide a more complete and realistic simulation framework for engineering applications, such as mission planning in GPS-denied environments, as well as for studying, monitoring or predicting the migration paths of animals that rely on magnetoreception.

\clearpage
% \bibliographystyle{jphysicsb}
% \bibliography{references}
\printbibliography

\end{document}